# Automated Human Activity Recognition by Colliding Bodies Optimization (CBO) -based Optimal Feature Selection with RNN

Pankaj Khatiwada[1]
Norwegian University of Science and Technology
Teknologivegen 22, 2815 Gjøvik, Norway
pankaj.khatiwada@ntnu.no

Ayan Chatterjee[2]*
University of Agder
Jon Lilletuns Vei 9, 4879 Grimstad, Norway
ayan.chatterjee@uia.no

Matrika Subedi[2]
University of Agder
Jon Lilletuns Vei 9, 4879 Grimstad, Norway
matris15@uia.no

## ABSTRACT

In an intelligent healthcare system, *Human Activity Recognition (HAR)* is considered an efficient approach in pervasive computing from activity sensor readings. The *Ambient Assisted Living (AAL)* in the home or community helps people provide independent care and enhanced living quality. However, many AAL models are restricted to multiple factors that include computational cost and system complexity. Moreover, the HAR concept has more relevance because of its applications, such as content-based video search, sports play analysis, crowd behavior prediction systems, patient monitoring systems, and surveillance systems. This study implements the HAR system using a popular deep learning algorithm, namely *Recurrent Neural Network (RNN)* with the activity data collected from intelligent activity sensors over time. The activity data is available in the *UC Irvine Machine Learning Repository (UCI)*. The proposed model involves three processes: (1) data collection, (b) optimal feature learning, and (c) activity recognition. The data gathered from the benchmark repository was initially subjected to optimal feature selection that helped to select the most significant features. The proposed optimal feature selection method is based on a new meta-heuristic algorithm called *Colliding Bodies Optimization (CBO)*. An objective function derived from the recognition accuracy has been used for accomplishing the optimal feature selection. The proposed model on the concerned benchmark dataset outperformed the conventional models with enhanced performance.

## CCS Concepts
• **Information systems** → **Information Retrieval**   • **Computing methodologies** → **Artificial Intelligence.**

## Keywords
Human Activity Recognition; Smart Activity Sensors; Optimal Feature Selection; Colliding Bodies Optimization; Recurrent Neural Network.

## 1. INTRODUCTION

Human beings can perform numerous activities concurrently, such as walking, communicating, eating and many more. HAR identifies the goals and actions of humans from a series of observations on the actions of humans. The daily life activities of HAR include sitting, sleeping, walking, standing, etc. Various deep learning models, such as DBN, CNN, RNN, ANN, DNN, CRNN, etc., and machine learning algorithms such as SVM, NB, KNN, etc., are used in previous works. For identifying different levels of activities, such as walking, moving upstairs or downstairs, sitting, standing, lying, and running, machine learning and deep learning algorithms can be applied.

The rapid growth of the Internet of things (IoT) has an advanced remote collection of human activity signals with the help of Wireless Sensor Network (WSN) [1][2]. The primary focus of our proposed activity recognition model is to detect the unexpected change in measures, such as covariance and mean that denotes the difference in an indoor environment [2]. The accurate manipulation of these measures is done using a robust algorithm. Activity recognition is relevant to the smart home model. For helping different emergency-related wellbeing services, such as fall detection for the elderly in healthcare, HAR is very helpful for aiding. Several physical activities are obtained for the nursing services and real-time responders in care homes and the domestic environment [1][3]. Introducing a scalable, robust, real-time indoor HAR model is a difficult task because of the complexity. The new technological developments and advancements stimulate the requirement for associated smart home sensing models and services [4]. In previous studies, different discrete classification models, such as Support Vector Machine (SVM), Naive Bayes (NB), K-Nearest Neighbor (KNN), Random Forest (RF), Conditional random fields (CRF), and Hidden Markov Model (HMM) have been used for human activity classifications, but deep learning algorithms can produce the better results as compared to the ML algorithms [5]. Some well-accepted deep learning algorithms used for HAR are Convolutional Neural Network (CNN), Long short-term memory (LSTM) and Recurrent Neural Network (RNN). To address the HAR and human pose recognition task, CNN is used in most applications by convoluting across two or three dimensions to seize a signal's spatial patterns [6].

HAR has become a lively and challenging research field in the past few years because of its applicability to various AAL domains to improve the demand for convenience services and home automation [7,12]. The camera usage is acceptable in most of the AAL scenarios. It provides various benefits in helping people with cognitive impairment, from event detection to person-environment interaction, assistive robots, and affective computing. The most typical applications of AAL are physiological monitoring, gait analysis, telerehabilitation, fall detection, and human behavior analysis. HAR has attained more interest in the ALL techniques in smart homes owing to the rapid increment of the world's aging population [7]. It is challenging to consider more count of observations in each second, the temporal nature of the observations, and the shortage of a straightforward procedure for relating accelerometer data for known movements [8]. Based on constant-size windows and training machine learning algorithms, such as for ensembles of decision trees, conventional algorithms consist of handcrafting features from the time-series data. In this area, feature engineering is a complex task, which necessitates deep proficiency [9] [10].

The contribution of the entire paper is highlighted below:

a. Introduction of an effective HAR model using RNN with proposed CBO algorithm-based optimal feature selection.
b. Recognizing human activities efficiently after deriving an objective function with maximum accuracy on two benchmark datasets, such as HAR dataset and WISDM dataset.

Regarding the paper's organization, Section 2 shows the literature review of existing HAR models, and Section 3 describes the HAR using sensed data. Section 4 speaks about the steps utilized for the proposed HAR, and Section 5 discusses the experimental analysis, and the paper is concluded in Section 6.

## 2. Related Work

This proposed work aims to develop an efficient HAR model with the help of RNN. HAR solves the problem of sequence classification of accelerometer data stored by smartphones or specialized harnesses into well-known movements. However, it has various drawbacks, such as partial occlusion, viewpoint, appearance, background clutter, scale variations, and lighting. Few related features and challenges are listed in Table 1.

Alhameed et al. [11] introduced an ambient HAR model using a multivariate Gaussian approach. The classification model has augmented prior information from passive Radio-frequency identification (RFID) tags to acquire more activity profiling. Based on multivariate Gaussian, the developed model with maximum likelihood estimation was employed for feature learning. In the mock apartment environment, twelve sequential and concurrent experimental analysis were performed. The developed model was well suitable for a single and multi-dwelling environment. Lizarazo et al. [12] performed the classification of six human activities with bidirectional LSTM networks, which employed IMF representation of inertial signals. From the UCI repository, the inertial signals of 2.56s records were gathered from 30 subjects with the help of a smartphone. By using ICEEMDAN, inertial signals were assembled for considering them to a similar scale and were combined with the IMF. The overall accuracy was acquired in classifying the six human activities. The results have proven that the developed ICEEMDAN was superior to conventional algorithms. Bidirectional LSTM [12] enhanced the observation capabilities of the system. The enhanced total ensemble empirical mode decomposition also provides a creative environment. Still, it seems very tough for inference as well as learning. Bernardini et al. [13] recommended various deep learning algorithms, which learned the human activities for classification. To model Spatio-temporal sequences, LSTM was implemented. The suggested model was analyzed using the "Center for Advanced Studies in Adaptive Systems" dataset. The results have been indicated that LSTM-based models were superior to conventional deep learning and machine learning algorithms. LSTM [13] does not require data augmentation methods, and it feels compatible with the highly unbalanced arrangement of the smart home dataset. It also enhances the generalization performance of the system. Still, the multi-user activity recognition was not implemented, and it also does not test several identical datasets. Fong et al. [14,3] have suggested a new data fusion model for merging data gathered from two sensors to improve the accuracy of HAR. To seize the complete details of body movements, Kinect can, but the accuracy was based on the angle of view. For detecting essential signals, wearable sensors were primitive in collecting spatial information but reliable. The combination of data from the two types of sensors has enabled each other by their strength. A new technique with incremental learning using the decision table was combined with swarm-based feature selection that was introduced for acquiring fast and precise HAR. The test results have been shown that HAR accuracy was improved when a wearing sensor was utilized simultaneously. Fast incremental learning [14] enhances the classification performance by the empirical data feeds, and it enhances the performance of the learning approach to around five times. However, it does not benchmark the performance using several machine learning, and various angles of the field of views from aerial are not included. It also does not extend the testing scenarios. Janko et al. [15] have suggested the analysis of the efficiency of components. For training conventional machine learning algorithms, complex feature selection and extraction approaches were utilized. By using end-to-end architecture, the training of deep learning models was done for combining deep multimodal Spectro-temporal. All the methods were combined to form the ensemble model, with last predictions smoothed using HMM for the activities of temporal dependencies. With the help of test data, the developed model has attained a more F1 score. The results have shown that the developed model has achieved the best HAR accuracy. Classical machine learning [15,4] optimizes the energy consumption with the help of sensor settings, and the activities are also predicted on an unlabeled dataset. However, with the recent techniques like multimodal subspace clustering or two-stream network fusion, it does not get updated, and with different subjects and datasets for AR, the model was not evaluated. These challenges are inspired to find a novel method for recognizing or predicting human activity using machine learning with data from smart sensors.

**Table 1. Review of Traditional Human Activity Prediction or Recognition models**

| Researcher | Method | Features | Challenges |
|---|---|---|---|
| Alhameed et al. [11] | Multivariate Gaussian | It attained a detailed description of activity profiling. It is applicable for single as well as multi-dwelling environments. It provides an extensive sensing environment for the disabled, elderly, and carers. | It becomes complicated to recognize complex activities. |
| Lizarazo et al. [12] | Bidirectional LSTM | It is creative with respect to the usage of the enhanced total ensemble empirical mode decomposition. It enhances the system's observability. | The inference and learning seem to be very tough. |
| Fong et al. [14] | Fast incremen | The performance of the machine learning | The testing scenarios are |

| | tal learning | models was increased to five times. The empirical data feeds to enhance the performance of the classification model. | not extended. It does not include various angles of the field of view from aerial. The performance was not benchmarked using several machine learning. |
|---|---|---|---|
| Janko et al. [15] | Classical machine learning | The activities are predicted on an unlabelled dataset. The sensor settings optimize energy consumption. | The model was not evaluated on different subjects as well as datasets for AR. It does not update with recent techniques such as multimodal subspace clustering or two-stream network fusion. |
| Bernardini et al. [13] | LSTM | The generalization performance is enhanced. It is compatible with the highly unbalanced arrangement of the smart home dataset. It does not need data augmentation methods. | Several identical datasets were not tested. It does not implement multi-user activity recognition. |

## 3. Method

This section describes the datasets that are gathered for human activity recognition and the proposed HAR model. Here two datasets are known as the HAR, and the WISDM dataset is used for the proposed HAR model. The RNN is used to classify the recognized activity for maximizing the accuracy with the help of the proposed CBO.

### 3.1 Datasets

Here, two datasets, namely HAR and WISDM, are utilized for analysis, and these datasets are downloaded from the UCI repository. Since these datasets are publicly available, they are considered here for better performance.

*HAR dataset:* This dataset is collected from the link [16] and the tests have been performed with 30 volunteers of age between 19 and 48 years (both male and female). Each person has performed six activities, such as walking, walking upstairs, walking downstairs, sitting, standing, and laying who is wearing a smartphone on the waist. To label the data manually, the simulations have been recorded in the video format. The acquired datasets are split into 70% of the data is considered training, and 30% for testing. The feature vector is acquired from each window by calculating the variables from the frequency and time domain. The sensor signals such as gyroscope and accelerometer are pre-processed using the noise filters.

*WISDM dataset:* This dataset is gathered from the link [17] and it is composed of the data gathered via the laboratory, controlled conditions with activities such as "walking, jogging, upstairs, downstairs, sitting, and standing." The collection of gyroscope and accelerometer sensor data is done from smartwatch and smartphone at a rate of 20Hz. This is acquired from 51 test subjects as they conduct 18 activities for 3 minutes apiece. In a discrete directory, the sensor data for each device and the sensor type is maintained. There are 51 files with respect to 51 test subjects in each directory. The entry format of each data is similar.

### 3.2 Proposed Model

In the earlier contributions, HAR is enabled in some applications, such as healthcare, manufacturing, and the smart homes. Activity recognition is essential for handling recorded data, thus permits the computing models for monitoring, analyzing, and assisting their daily life. HAR recognizes the particular action or movement of a person based on the sensor data. In the case of activity recognition, sensor data seems to be an expensive and challenging one that needs custom hardware. Learning methods can be used for automatically learning the features for making accurate predictions from the raw data in a direct manner. This can lead to new sensor modalities, new datasets, and new problems that can be adopted in a cheap and fast manner. The usage of smart sensors in the current healthcare models helps health professionals and patients to automatically monitoring human activities. In personal healthcare monitoring, smartphones and smart body sensors are rapidly employed. The wearable sensor technology is a significant improvement in smart sensor technologies. Moreover, there is more interest in machine learning algorithms, and it plays a significant role in HAR. The proposed HAR model is represented in Figure 1. In the proposed HAR model, the two datasets, such as HAR and WISDM, are collected from the UCI public repository. The developed model includes the following steps - data collection, optimal feature selection, and activity recognition. Based on the developed CBO algorithm, the optimal feature selection was performed. For effective Recognition of human activities, a deep learning model namely RNN has been used. The main objective of the proposed HAR model is to maximize classification accuracy.

### 3.3 Steps Utilized for Proposed Human

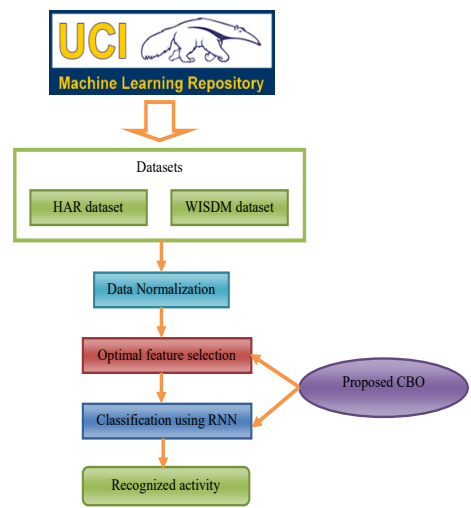

**Figure 1. The proposed human activity recognition model.**

## Activity Recognition

The selected principal features for the activity recognition are - data normalization, optimal feature selection by colliding bodies optimization, and RNN-based activity recognition.

### 3.3.1 Data Normalization

Data normalization is a process in which the data in the database is appropriately arranged, which is used for altering the numerical values in the dataset to a common scale without distorting the ranges. By the frequency of occurrence, the data normalization procedure is defined. With this concept, the record-level normalization produces the data representation related to the record, and it is generally observed between the same record set for an entity. In the normalized record, the field level normalization selects the value for each field, which occurs often. Since both the gyroscopic and accelerometer data are combined, the gravitational effect is removed. Let $am$ and $bm$ be the two variables, where the maximum and minimum normalized values are indicated by $am$ and $bm$, respectively. The data normalization is mathematically represented in Eq. (1)

$$DT_u^{norm} = (am - bm) \times \frac{(DT_u - DT^{min})}{(DT^{max} - DT^{min})} + bm \tag{1}$$

In the above equation, the term $DT_u^{norm}$ represents the normalized data, and the term $DT_u$ denotes the value or data that to be normalized. The maximum and minimum values related to each record is denoted as $DT^{max}$ and $DT^{min}$, correspondingly.

### 3.3.2 Optimal Feature Selection by Colliding Bodies Optimization

From the normalized data, the optimal features are selected. The employment of optimal feature selection is to decrease the dimensionality of the data for developing the best classification model. The results of classification are impacted using optimal feature selection approaches. If optimal feature selection is good, there is a positive effect on classification in the proposed HAR. With the help of developed CBO-RNN, the optimal feature selection is performed. The solution encoding of optimal feature selection is given in Fig. 2

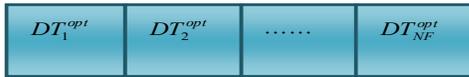

Fig. 2. Solution encoding of proposed human activity recognition

From the above figure, the optimal features are denoted as $DT_u^{opt}$, in which the number of features is indicated by $NF$. The objective function of the proposed HAR is to maximize the accuracy, which is attained by the proposed CBO-RNN. The objective function of the proposed model is denoted in Eq. (2).

$$obj = \arg\max_{\{DT_u^{norm}\}}(acc) \tag{2}$$

The numerical formula for computing the accuracy is specified in Eq. (3).

$$A = \frac{Tps + Fps}{Tps + Fps + Tne + Fne} \tag{3}$$

In the above equation, the term $TP_s$ denotes true positive of the elements, $T_{ne}$ denotes true negative, $FP_s$ indicates false positive, and $Fne$ indicates the false negative. CBO [18] is a population-based evolutionary algorithm that uses the concept of the laws of the collision of two objects. To find a maximum or minimum of functions, CBO employs simple formulation and does not depend on any internal parameter. Each solution candidate $x_i$ includes the count of variables, and it is assumed as the CB. The massed objects include stationary and moving objects, in which the stationary objects are followed by moving objects, and the collision will occur among the object pairs. It is performed for enhancing the moving object's position and pushing stationary objects towards the best position. Based on new velocities, the novel positions of CBs are updated using collision laws. The process of conventional CBO is given below:

**A.** By initializing the population in the search space at random, the initial positions of CBs are defined as shown in Eq. (4). Here, the initial value vector of $i^{th}$ CB is given by $X_i^0$. The minimum and maximum allowable values vectors of variables are denoted as $X_{min}$ and $X_{max}$, respectively. Moreover, the random number is given by $rnd$, which lies from 0 to 1, and the count of CBs is given by $N$.

$$X_i^0 = X_{min} + rnd(X_{max} - X_{min}), i = 1, 2, \cdots, N \tag{4}$$

**B.** For each CB, the magnitude of the body mass is given by Eq. (5). In this, the objective function value of the agent $i$ is denoted as $ft(b)$, and the population size is denoted as $N$. The objective function $ft(b)$ is replaced by $1/ft(b)$ for maximization.

$$mg_b = \frac{\frac{1}{ft(b)}}{\sum_{i=1}^{N} \frac{1}{ft(b)}}, \quad b = 1, 2, \cdots, N \tag{5}$$

**C.** The CB's objective values are allotted in ascending order. The sorted CBs are similarly split into two groups. The lower half CBs are termed as stationary CBs. These CBs are good agents that are stationary, and these bodies velocity is 0 before collision based on Eq. (6).

$$vl_i = 0, i = 1, 2, \cdots, \frac{N}{2} \tag{6}$$

The upper half CBs are called as moving CBs, which will move to the lower half CBs. The velocity of these bodies is given by the position of the body prior to the collision, as shown in Eq. (7).

$$vl_i = X_i - X_{i-\frac{N}{2}}, i = \frac{N}{2} + 1, \cdots, N \tag{7}$$

In the above equation, the velocity and position vector of $i^{th}$ CB is given by $vl_i$ and $X_i$, respectively. The $i^{th}$ CB pair position of $X_i$, in the last group is given by $X_{i-\frac{N}{2}}$.

**D.** In each group, the velocities of CBs are analysed using Eq. (6) and Eq. (7), and the velocity before the collision. By using Eq. (8), the velocity of each moving CBs after the collision is acquired.

$$vl'_i = \frac{\left(mg_i - \varepsilon mg_{i-\frac{N}{2}}\right)vl_i}{mg_i + mg_{i-\frac{N}{2}}}, \quad i = \frac{N}{2}+1,\cdots,N \quad (8)$$

In Eq. (8), the velocity of $i^{th}$ moving CB before and after the collision is given by $vl_i$ and $vl'_i$, respectively. The $i^{th}$ CBs mass is given by $mg_i$. The mass of $i^{th}$ CB pair is denoted as $mg_{i-\frac{N}{2}}$. Once the collision is done, the velocity of each stationary CB is denoted in Eq. (9).

$$vl'_i = \frac{\left(mg_i + \varepsilon mg_{i+\frac{N}{2}}\right)vl_{i+\frac{N}{2}}}{mg_i + mg_{i+\frac{N}{2}}}, \quad i = \frac{N}{2}+1,\cdots,N \quad (9)$$

In the above equation, the moving CB pair before and stationary CB after collision's velocity is given by $vl_{i+\frac{N}{2}}$ and $vl'_i$, respectively. The value of the COR parameter is denoted as $\varepsilon$.

**E.** By using the generated velocities after collision in the stationary CB's position, the new locations of CBs are analysed. Eq. (10) represents the new locations of each moving CBs. Here, the $i^{th}$ moving CBs new position and velocity after a collision is denoted as $X_i^{new}$ and $vl'_i$, respectively. The old position of $i^{th}$ stationary CB pair is given by $X_{i-\frac{N}{2}}$. The new locations of stationary CBs are acquired using Eq. (11).

$$X_i^{new} = X_{i-\frac{N}{2}} + rnd \circ vl'_i, \quad i = \frac{N}{2}+1,\cdots,N \quad (10)$$

$$X_i^{new} = X_i + rnd \circ vl'_i, \quad i = 1,\cdots,\frac{N}{2} \quad (11)$$

In Eq. (8) the random vector is uniformly distributed, and the value ranges from -1 to 1, and the element-by-element multiplication is given by $\circ$.

**F.** Until the termination criterion is reached, the optimization procedure is repeated from step 2. Termination criteria mean the count of maximum iteration is fulfilled. The status of the body and its numbering is modified in two successive iterations.

### 3.3.3 RNN-based Recognition

RNN [19] is a dynamic model, which is computationally powerful, and it is used in many temporal processing methods and applications. The model is trained for providing any target dynamics until the degree of precision is offered. RNN is one of the categories of ANNs, where the links among the nodes generate the directed graph with the information growth. This exhibits temporal dynamic behaviour. It uses their memory or the internal state for processing the input's variable-length sequences. RNN describes two broad classes of networks having an identical general structure, such as finite and infinite impulse. The first one is a directed acyclic graph that can be unrolled and replaced using a strictly feedforward NN, and the second one is a directed cyclic graph, and it cannot be unrolled. LSTM is one type of RNN that consists of three gate units, such as input, output, and forget gates, and a memory cell unit. The LSTM is subjected to long-term dependencies like anomaly detection, speech detection, handwriting recognition, forecasting, and time-series in the network traffic. These are mostly used in deep learning. LSTM is composed of a chain-like structure consisting of a forget gate, an output gate, an input gate, and a cell. LSTM involves feedback connections. It performs a better classification process, and the predictions are accurate based on the time-series data. It overcomes the problem like vanishing gradients and exploding gradients. The distinct LSTM model types are the multivariate multi-step, multi-step, multivariate, and univariate used for the time-series forecasting problem. The several steps for training and testing an LSTM include importing the python libraries, loading the data from the repository, data pre-processing, feature scaling, splitting of the univariate sequence into samples, creating LSTM models, and evaluating the performance etc. The specific type of LSTM called GRU [20] is considered, which is employed for building the model of RNN for enhancing the performance. It combines both forget and output gates into one update gate $U_{p_a}$, where the interpolation is used for attaining the current result. Assume $g_a \leftarrow DT_u^{opt}$ as the $a^{th}$ input feature and the earlier hidden state is denoted as $d_{a-1}$. The update and reset gates are denoted in Eq. (12) and Eq. (13), respectively. Here, the activation function is denoted as $A_{cv}$, which is the logistic sigmoid function.

$$Up_a = Acv\left(wm^{dUp}d_a + wm^{dUp}d_{a-1}\right) \quad (12)$$

$$Rgt_a = Acv\left(wm^{dRgt}g_a + wm^{dRgt}d_{a-1}\right) \quad (13)$$

In the above equations, the weight matrix is given by $wm^a = \left\{wm^{dUp}, wm^{dUp}, wm^{gRgt}, wm^{dRgt}\right\}$ that must be tuned for error minimization among actual and measured output. The RNN is diagrammatically showed in Figure 2.

The hidden unit's candidate state is measured by Eq. (14).

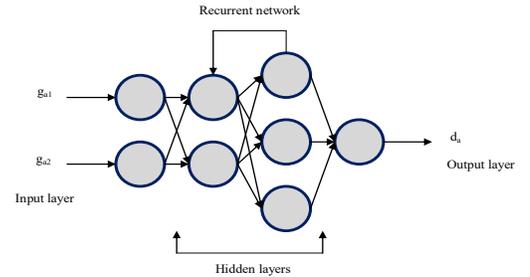

**Figure 2. Diagrammatic representation of RNN.**

$$\tilde{d}_a = \tan\left(wm^{gd}e_d + wm^{dd}\left(d_{a-1} \otimes Rgt_a\right)\right) \quad (14)$$

In Eq. (14) The term $\otimes$ represents the element-wise multiplication; the hidden activation function of the candidate state is denoted as $\tilde{d}_a$, and the linear interpolation $d_{a-1}$ is given by $d_a$ of GRU, and the numerical equation is given in Eq. (15).

$$d_a = (1-Up_a) \otimes \tilde{d}_a + Up_a \otimes d_{a-1} d_a = (1-Up_a) \otimes \tilde{d}_a + Up_a \otimes d_{a-1} \quad (15)$$

In the proposed HAR, the objective is to maximize the recognition accuracy with the optimal features.

## 4. Results and Discussion

### 4.1 Experimental Setup
The presented HAR was implemented using PyCharm, and the analysis was carried out. The datasets named UCI-HAR and WISDM were considered for the experiment. The maximum number of iterations considered for the experiment was 25, and the population size was considered as 10. The analysis of the proposed CBO-RNN was compared over conventional meta-heuristic algorithms like PSO-RNN [21], FF-RNN [22], and CBO-RNN [18], and the performance of RNN was compared over NN [23], DT [24], KNN [25], SVM [26] concerning the performance metrics, such as accuracy, sensitivity, specificity, precision, FPR, FNR, NPV, FDR, F1 score, and MCC.

### 4.2 Performance Measurement
The ten performance measures used are described below:

Accuracy: It is specified in Eq. (3).

Sensitivity: It is the number of true positives, which are recognized exactly.

$$Sensi = \frac{Tps}{Tps + Fne} \quad (16)$$

Specificity: It is the number of true negatives, which are determined precisely.

$$Speci = \frac{Tne}{Fps} \quad (17)$$

Precision: It is, the ratio of positive observations that are predicted exactly to the total number of observations that are positively predicted.

$$Preci = \frac{Tps}{Tps + Fps} \quad (18)$$

FPR: It is the ratio of the count of false-positive predictions to the entire count of negative predictions.

$$FPR = \frac{Fps}{Fps + Tne} \quad (19)$$

FNR: It is the proportion of positives that yield negative test outcomes with the test.

$$FNR = \frac{Fne}{Tne + Tps} \quad (20)$$

NPV: It is the probability that subjects with a negative screening test truly do not have the disease.

$$NPV = \frac{Fne}{Fne + Tne} \quad (21)$$

FDR: It is the number of false positives in all the rejected hypotheses.

$$FDR = \frac{Fps}{Fps + Tps} \quad (22)$$

F1 Score: It is the harmonic mean between precision and recall, and it is used as a statistical measure to rate performance.

$$F1\,Sco = \frac{Sensi \bullet Preci}{Preci + Sensi} \quad (23)$$

MCC: It is, the correlation coefficient is computed by four values.

$$MCC = \frac{Tps \times Tne - Fps \times Fne}{\sqrt{(Tps+Fps)(Tps+Fne)(Tne+Fps)(Tne+Fne)}} \quad (24)$$

### 4.3 Analysis of Diverse heuristic-based Optimal Feature Selection
The analysis of the proposed and the traditional meta-heuristic algorithms for optimal feature selection concerning learning percentage for UCI-HAR and WISDM datasets is shown in Figure 3. In Figure 3 (a), the accuracy of the developed CBO-RNN is acquiring the best results in recognizing the human activities compared to conventional algorithms for the UCI-HAR dataset. The accuracy of the improved CBO-RNN at learning percentage 85 is 2.2% better than PSO-RNN and 3.4% superior to FF-RNN. Table 2 and Table 3 show the overall analysis of the developed CBO-based optimal feature selection over the traditional algorithms for UCI-HAR and WISDM datasets through RNN-based classification. In Table 2, the accuracy of the implemented CBO-RNN is 1.4% improved than PSO-RNN and 2.6% improved than FF-RNN. From Table 3, the accuracy of the suggested CBO-RNN is 1.4% better than PSO-RNN, 1.7% better than FF-RNN. Thus, the developed CBO-based optimal feature selection has attained the best results in recognizing human activities.

### 4.4 Analysis of Various Classifiers
In Figure 4, the analysis of various classifiers concerning learning percentage using UCI-HAR and WISDM datasets is shown. The optimal features by CBO are used for performing the Recognition by all classifiers. For the WISDM dataset, the accuracy of the presented RNN when considering the learning percentage as 75 is 2.1% advanced than NN, 5.5% advanced than SVM and KNN, and 9.1% advanced than DT. The overall performance analysis of the developed and the conventional classifiers using two datasets is shown in Table 4 and Table 5. In Table 4, the accuracy of the proffered RNN is 7.9% improved than DT, 6.4% advanced than KNN, 5.7% advanced than SVM, and 2.4% advanced than NN using UCI-HAR dataset. Table 5 shows the overall performance analysis of the proposed CBO-RNN and the conventional classifiers for the WISDM dataset. The accuracy of the presented RNN is attaining the best HAR. It is 7.3% progressed than DT, 6.5% advanced than KNN, 4.1% advanced than SVM, and 2.2% advanced than NN. Hence, it is confirmed that the suggested RNN is performing well in recognizing human activities with CBO-based optimal features.

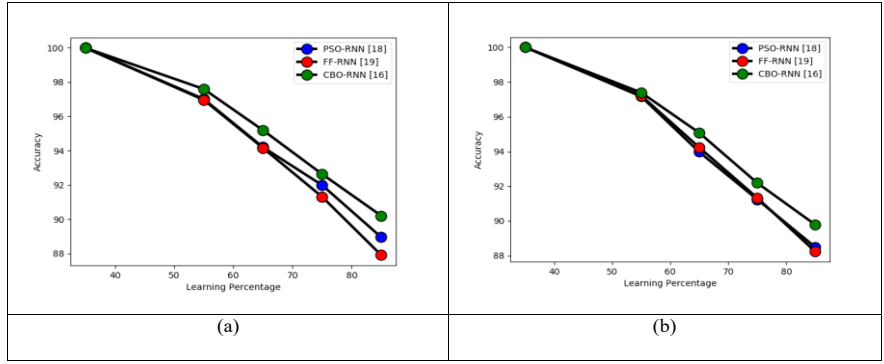

|     (a)     |     (b)     |

**Figure 3. Analysis of Diverse heuristic-based Optimal Feature Selection for HAR concerning Accuracy using (a) UCI-HAR dataset, and (b) WISDM.**

**Table 2. Different Heuristic-based optimal feature selection for HAR using the UCI-HAR dataset**

| Algorithms | Accuracy | Sensitivity | Specificity | Precision | FPR | FNR | NPV | FDR | F1 score | MCC |
|---|---|---|---|---|---|---|---|---|---|---|
| PSO-RNN | 0.889443 | 0.8902 | 0.889292 | 0.616593 | 0.110708 | 0.1098 | 0.889292 | 0.383407 | 0.728556 | 0.679592 |
| FF-RNN | 0.879008 | 0.888385 | 0.877132 | 0.591184 | **0.122868** | **0.111615** | 0.877132 | **0.408816** | 0.709935 | 0.658454 |
| CBO-RNN | **0.901996** | **0.909256** | **0.900544** | **0.646452** | 0.099456 | 0.090744 | **0.900544** | 0.353548 | **0.755656** | **0.71239** |

**Table 3. Different Heuristic-Based optimal feature selection for HAR using the WISDM dataset.**

| Algorithms | Accuracy | Sensitivity | Specificity | Precision | FPR | "FNR" | NPV | FDR | F1 score | MCC |
|---|---|---|---|---|---|---|---|---|---|---|
| PSO-RNN | 0.884846 | **0.899563** | 0.88398 | 0.313229 | 0.11602 | 0.100437 | 0.88398 | 0.686771 | 0.464662 | 0.490129 |
| FF-RNN | 0.882258 | 0.870451 | 0.882952 | 0.304326 | **0.117048** | **0.129549** | 0.882952 | **0.695674** | 0.45098 | 0.472052 |
| CBO-RNN | **0.897784** | 0.887918 | **0.898365** | **0.339455** | 0.101635 | 0.112082 | **0.898365** | 0.660545 | **0.491143** | **0.511057** |

## 4.5 Effect of Optimal Feature Selection

The effect of optimal feature selection on both UCI-HAR and WISDM is shown in Figure 5. Figure 5 (a) shows that the accuracy of the optimized feature at learning percentage 85 is 2.2% superior to all features. Moreover, the overall analysis of with and without optimized features for both the datasets is shown in Table 6 and Table 7. From Table 6, the accuracy of the optimized feature is 1.8% enhanced than all features. In Table 7, the accuracy of the optimized features is attaining the best results for HAR. It is 1.5% better than all features. Therefore, it can be concluded that the developed CBO-RNN has generated better results in recognizing human activities.

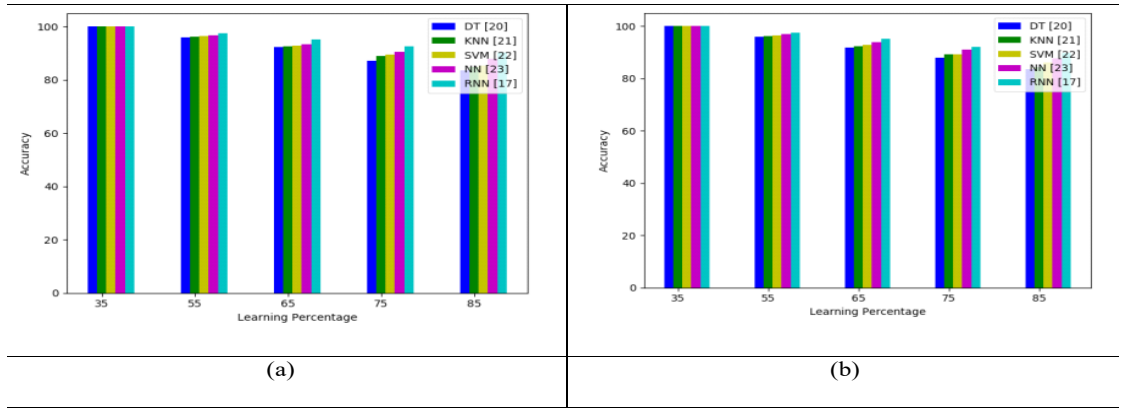

| (a) | (b) |
|---|---|

Figure 4. Analysis on diverse machine learning algorithms for HAR concerning Accuracy using (a) UCI-HAR dataset, and (b) WISDM.

Table 4. Different Machine Learning algorithms for HAR using the UCI-HAR dataset.

| Algorithms | Accuracy | Sensitivity | Specificity | Precision | FPR | FNR | NPV | FDR | F1 score | MCC |
|---|---|---|---|---|---|---|---|---|---|---|
| **DT [20]** | 0.835451 | 0.842105 | 0.83412 | 0.5038 | 0.16588 | **0.157895** | 0.83412 | **0.4962** | 0.630435 | 0.562152 |
| **KNN [21]** | 0.847701 | 0.850272 | 0.847187 | 0.5267 | 0.152813 | 0.149728 | 0.847187 | 0.4733 | 0.650469 | 0.586123 |
| **SVM [22]** | 0.853297 | 0.856624 | 0.852632 | 0.537585 | **0.147368** | 0.143376 | 0.852632 | 0.462415 | 0.660602 | 0.598505 |
| **NN [23]** | 0.880067 | 0.869328 | 0.882214 | 0.596142 | 0.117786 | 0.130672 | 0.882214 | 0.403858 | 0.707272 | 0.652996 |
| **RNN [17]** | **0.901996** | **0.909256** | **0.900544** | 0.646452 | 0.099456 | 0.090744 | **0.900544** | 0.353548 | **0.755656** | **0.71239** |

Table 5. Different Machine Learning algorithms for HAR using the WISDM dataset.

| Algorithms | Accuracy | Sensitivity | Specificity | Precision | FPR | FNR | NPV | FDR | F1 score | MCC |
|---|---|---|---|---|---|---|---|---|---|---|
| **DT [20]** | 0.836568 | 0.825328 | 0.837229 | 0.229741 | **0.162771** | **0.174672** | 0.837229 | **0.770259** | 0.359429 | 0.379715 |
| **KNN [21]** | 0.842956 | 0.845706 | 0.842795 | 0.240381 | 0.157205 | 0.154294 | 0.842795 | 0.759619 | 0.374356 | 0.397702 |
| **SVM [22]** | 0.862203 | 0.863173 | 0.862146 | 0.269178 | 0.137854 | 0.136827 | 0.862146 | 0.730822 | 0.410381 | 0.434202 |
| **NN [23]** | 0.877972 | 0.861718 | 0.878928 | 0.295115 | 0.121072 | 0.138282 | 0.878928 | 0.704885 | 0.439658 | 0.4602 |
| **RNN [17]** | **0.897784** | **0.887918** | **0.898365** | **0.339455** | 0.101635 | 0.112082 | **0.898365** | 0.660545 | **0.491143** | **0.511057** |

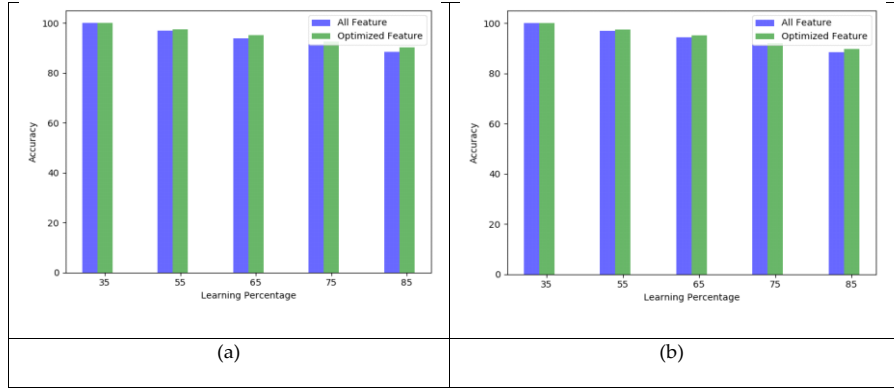

| (a) | (b) |

**Figure 5. Effect of optimal feature selection for HAR concerning Accuracy using (a) UCI-HAR dataset, and (b) WISDM.**

**Table 6. Overall analysis - with and without optimized features for HAR using UCI-HAR dataset**

| Algorithms | Accuracy | Sensitivity | Specificity | Precision | FPR | FNR | NPV | FDR | F1 score | MCC |
|---|---|---|---|---|---|---|---|---|---|---|
| All Features | 0.885209 | 0.878403 | 0.88657 | 0.607659 | **0.11343** | **0.121597** | 0.88657 | **0.392341** | 0.718367 | 0.666647 |
| Optimized Feature | **0.901996** | **0.909256** | **0.900544** | **0.646452** | 0.099456 | 0.090744 | **0.900544** | 0.353548 | **0.755656** | **0.71239** |

**Table 7. Overall analysis - with and without optimized features for HAR using the WISDM dataset.**

| Algorithms | Accuracy | Sensitivity | Specificity | Precision | FPR | FNR | NPV | FDR | F1 score | MCC |
|---|---|---|---|---|---|---|---|---|---|---|
| All features | 0.884441 | **0.89083** | 0.884065 | 0.311292 | **0.115935** | 0.10917 | 0.884065 | **0.688708** | 0.461364 | 0.485418 |
| Optimized Feature | **0.897784** | 0.887918 | **0.898365** | **0.339455** | 0.101635 | **0.112082** | **0.898365** | 0.660545 | **0.491143** | **0.511057** |

## 5. Conclusion

In this study, we have described how to design, develop a novel HAR model with deep learning algorithms and test it further on publicly available (UCI repository) HAR datasets. The initial CBO-based optimal feature selection method helped us to select the most relevant features. To accomplish the optimal feature selection, an objective function was derived with the help of recognition accuracy. Furthermore, RNN was used to recognize human activities. From this analysis, we obtained better accuracy from our developed CBO-RNN model to recognize human activities when compared over conventional algorithms for the UCI-HAR dataset. Therefore, we can conclude that the developed CBO-RNN is more efficient than others to recognize human activities under defined settings.